\documentclass[10pt]{article}

\usepackage[T1]{fontenc}
\usepackage[utf8]{inputenc}
\usepackage{lmodern}
\usepackage[margin=0.82in]{geometry}
\usepackage{microtype}

\usepackage{amsmath}
\usepackage{amssymb}
\usepackage{graphicx}
\usepackage{booktabs}
\usepackage{array}
\usepackage{nicefrac}
\usepackage{subcaption}
\usepackage{multirow}
\usepackage{siunitx}
\usepackage{colortbl}
\usepackage{longtable}
\usepackage{xcolor}
\usepackage[numbers,sort&compress]{natbib}
\usepackage[hidelinks]{hyperref}

\hypersetup{
  pdftitle={UAV3DCrop: Benchmarking 3D Reconstruction in Repeated Multi-Angle UAV Crop Surveys},
  pdfauthor={Junxiong Zhou et al.},
  pdfsubject={Preprint},
  pdfkeywords={UAV imagery, agricultural datasets, crop-field reconstruction, neural radiance fields, Gaussian splatting}
}

\providecommand{\Description}[1]{}

\definecolor{statsupportblue}{RGB}{226,238,247}
\newcommand{\tabbest}[1]{\textbf{#1}}
\newcommand{\tabsecond}[1]{\underline{#1}}
\newcommand{\tabsupported}[1]{\cellcolor{statsupportblue}\textbf{#1}}


\title{\textbf{UAV3DCrop: Benchmarking 3D Reconstruction in Repeated Multi-Angle UAV Crop Surveys}}

\author{%
Junxiong Zhou\textsuperscript{1,2,*,\(\dagger\)},
Xuechen Li\textsuperscript{1,*},
Chonghao Qiu\textsuperscript{3,*},
Lang Qiao\textsuperscript{1},
Xiaowei Jia\textsuperscript{3}\\[0.25em]
Qi Yang\textsuperscript{4},
Chishan Zhang\textsuperscript{5},
Leikun Yin\textsuperscript{1},
Nanshan You\textsuperscript{1},
Vipin Kumar\textsuperscript{1}\\[0.25em]
David Mulla\textsuperscript{1},
Ce Yang\textsuperscript{1},
Zhenong Jin\textsuperscript{1,6,\(\dagger\)},
Licheng Liu\textsuperscript{1,2,\(\dagger\)}
\\[0.8em]
\small\textsuperscript{1}University of Minnesota, Twin Cities, USA\\
\small\textsuperscript{2}University of Wisconsin--Madison, USA\\
\small\textsuperscript{3}University of Pittsburgh, USA\\
\small\textsuperscript{4}Max Planck Institute for Biogeochemistry, Germany\\
\small\textsuperscript{5}Boston University, USA\\
\small\textsuperscript{6}Peking University, China\\[0.4em]
\small\textsuperscript{*}Equal contribution.
\textsuperscript{\(\dagger\)}Correspondence:
\href{mailto:zhou1743@umn.edu}{zhou1743@umn.edu},
\href{mailto:jinzn@umn.edu}{jinzn@umn.edu},
\href{mailto:licheng.liu@wisc.edu}{licheng.liu@wisc.edu}
}

\date{\small Preprint}

\begin{document}

\maketitle

\begin{abstract}
Accurate 3D crop monitoring underpins data-driven precision agriculture by enabling field-scale analysis of plant structure, growth dynamics, and management response. Modern 3D reconstruction methods perform strongly on generic benchmarks, but rendered appearance may not translate into metrically and agronomically useful geometry in crop fields. We introduce UAV3DCrop, a public benchmark of repeated multi-angle unmanned aerial vehicle (UAV) crop surveys. It contains 88,830 RGB images at $5280 \times 3956$ pixels, with a ground sampling distance of 3.6--5.8~mm, from 91 scenes spanning corn, soybean, wheat, and oat. Track A evaluates seven scene-optimized methods---Neural Radiance Field (NeRF) and 3D Gaussian Splatting (3DGS) variants---on held-out views, photogrammetry-referenced depth, and canopy-height recovery. Track B tests four pretrained feed-forward models on zero-shot camera-pose and geometry estimation. The scene-optimized methods rank differently across the three targets: Splatfacto-big leads appearance, whereas Scaffold-GS leads depth and is statistically tied with Splatfacto for canopy height. Among feed-forward models, MapAnything leads on seven of the eight metrics, while the remaining models vary more across crops and fail severely on absolute scale in a way that alignment conceals. Repeated acquisitions reveal further sensitivities that differ by output type and by model, associated with position within the acquisition sequence and with tie-point multiplicity. Current 3D reconstruction methods are therefore not yet interchangeable for agronomic use: no single method wins on appearance, geometry, and canopy height at once, and only one of four feed-forward models recovers usable metric scale. The dataset is publicly available at \url{https://link-dev.github.io/UAV3DCrop/}.

\end{abstract}


\section{Introduction}
\label{sec:intro}

High-throughput crop phenotyping relies on repeated, fine-scale field observations to quantify crop growth and guide precision agriculture \cite{khanal2017overview,sishodia2020applications,xie2020review}. Unmanned aerial vehicle (UAV) sensing has expanded this capability, but many pipelines reduce overlapping images to 2D orthomosaics, vegetation indices, or image-level features \cite{yang2017unmanned,jin2020high}. These products serve classification, stress monitoring, and yield prediction well, but represent canopy geometry only indirectly, whereas canopy height, leaf distribution, and plot-level architecture are inherently 3D and change throughout crop development.

\begin{figure*}[t]
  \centering
  \includegraphics[width=1.0\linewidth]{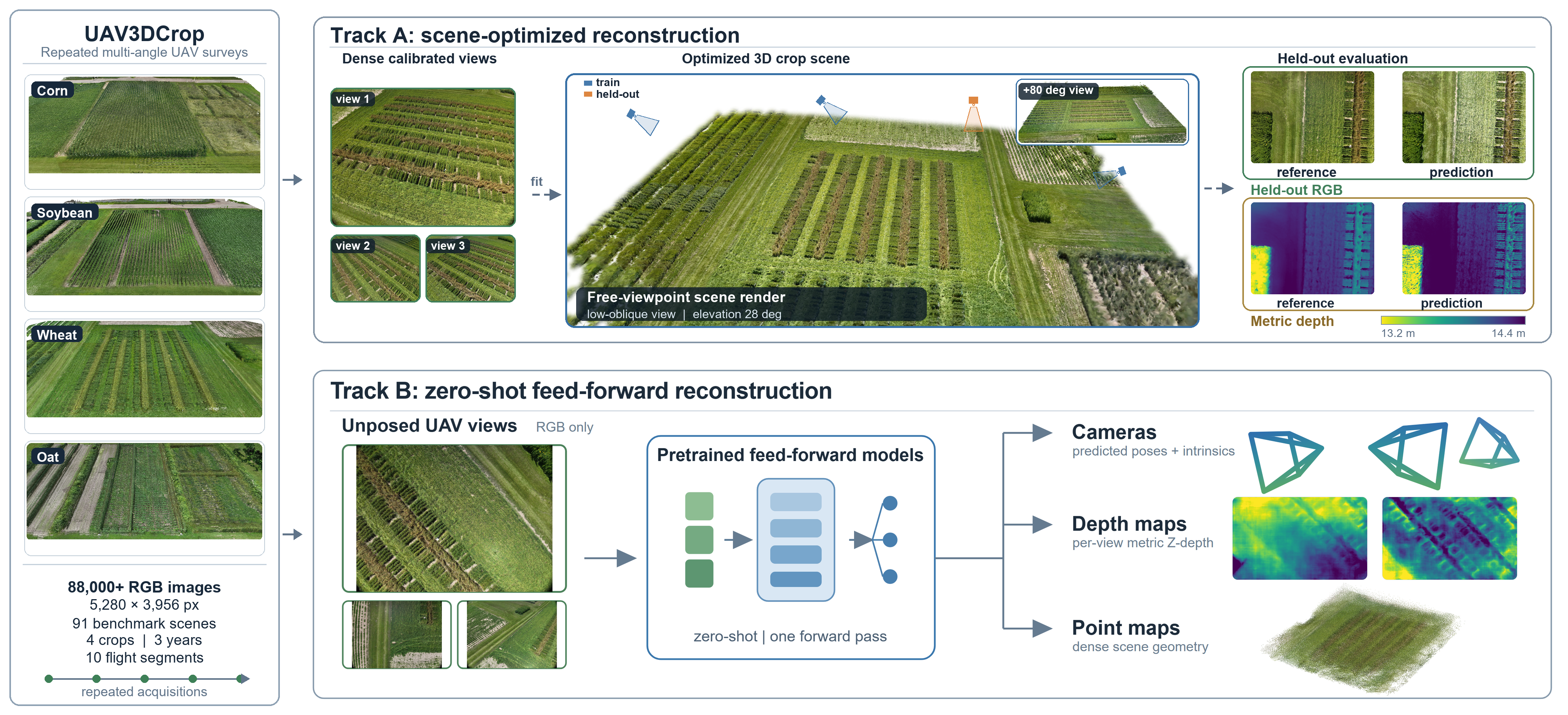}
  \caption{Overview of UAV3DCrop and its two-track benchmark. Track A evaluates scene-optimized reconstruction from dense posed views against held-out RGB and photogrammetry-referenced depth; Track B evaluates zero-shot feed-forward geometry from unposed views.}
  \Description{A dataset strip for corn, soybean, wheat, and oat connects to two benchmark pipelines: scene-optimized reconstruction with held-out appearance and depth evaluation, and zero-shot feed-forward reconstruction producing cameras, depth maps, and point maps.}
  \label{fig:fig1}
\end{figure*}

Scene-optimized neural rendering offers a route from posed multi-view UAV imagery to 3D crop representations. Neural Radiance Fields (NeRFs) and 3D Gaussian Splatting (3DGS) fit a separate representation to each scene and synthesize novel views \cite{mildenhall2021nerf,kerbl20233d,tancik2023nerfstudio}. Agricultural studies use these models for panoptic crop representation \cite{smitt2023pag}, boll mapping and plant architecture \cite{jiang2025cotton3dgaussians}, and wheat-head reconstruction \cite{zhang2025wheat3dgs}, but each covers a single crop, acquisition setting, or downstream task.

Crop canopies also differ from the scenes general 3D benchmarks sample. Rows repeat at near-constant spacing and leaves share color and texture, leaving feature matching little to anchor on; dense foliage occludes itself; thin leaves sit at the limit of what current representations resolve; and the canopy deforms between passes as plants move. Static single-date evaluation scenes such as those of ETH3D \cite{schops2017multi} and Mip-NeRF 360 \cite{barron2022mip} do not jointly target these crop-specific conditions.

Feed-forward visual-geometry models raise a complementary question: large-scale pretraining lets them predict cameras, depth, and point maps directly, without fitting each scene \cite{wang2024dust3r,wang2025vggt,keetha2025mapanything}. Requiring no per-scene optimization, they scale naturally to repeated surveys, and crop imagery is a demanding domain-shift test for models pretrained on general-purpose scenes.

We introduce UAV3DCrop, a benchmark of repeated multi-angle UAV crop surveys spanning 88,830 images, 91 scenes, four crops, and three seasons, with refined poses, a photogrammetric depth reference, and linked field measurements (Fig. \ref{fig:fig1}; Sec.~\ref{sec:dataset}). We organize it around three questions. \textbf{RQ1:} Can scene-optimized methods jointly recover high-fidelity appearance and reliable photogrammetry-referenced geometry in field-scale crop scenes? \textbf{RQ2:} Do standard appearance and geometry metrics agree with downstream agronomic utility, as measured by canopy-height recovery? \textbf{RQ3:} Can pretrained feed-forward models transfer zero-shot to crop imagery and recover absolute metric scale? Repeated acquisitions provide a cross-cutting stress dimension for testing the stability of these answers.

These questions do not have reassuring answers. Method rankings depend on the evaluation target: the strongest renderer is not the strongest reconstructor, and increasing model capacity can improve appearance while degrading geometry. Most feed-forward models recover accurate geometry only up to an unknown scale, so the alignment step that makes them look competitive also hides the failure that metric agronomic use would encounter first.

Our contributions are: (1) a public, field-scale UAV dataset with fixed manifests, quality-control (QC) metadata, and linked plant-height and effective leaf area index (LAI) measurements; (2) a standardized two-track benchmark and a systematic evaluation of seven scene-optimized methods and four zero-shot feed-forward models across appearance, photogrammetry-referenced geometry, efficiency, metric scale, and downstream canopy height; and (3) evidence that these targets induce different method rankings, which sets priorities for reliable field-scale crop phenotyping.

\section{Related Work}
\label{sec:related}

\noindent\textbf{UAV crop phenotyping and multi-view crop datasets.}
UAV remote sensing enables repeated crop observation at very high spatial and temporal resolution for high-throughput phenotyping \cite{xie2020review,sishodia2020applications}. Multi-view imagery and active sensors extend phenotyping to canopy height, organ distribution, and plant architecture \cite{zhu2023quantitative,xiao2023high,lin2021quality,rivera2023lidar}. GroMo25 records indoor growth \cite{bansal2025gromo25}, TomatoMAP targets fine-grained tomato phenotyping \cite{zhang2025tomato}, and MIPDB combines ground and UAV imagery for time-series maize analysis \cite{wang2024mipdb}. These resources address controlled growth, single-crop phenotyping, or ground--UAV time series rather than repeated multi-directional field reconstruction across crops.

\noindent\textbf{Scene-optimized reconstruction and neural rendering.}
Classical reconstruction estimates cameras with structure from motion (SfM) and dense geometry with multi-view stereo (MVS) \cite{snavely2006photo,schonberger2016structure,goesele2007multi}. NeRFs optimize continuous radiance fields \cite{mildenhall2021nerf}, whereas 3DGS uses explicit anisotropic Gaussians for efficient rendering \cite{kerbl20233d,wu2024recent}; both are scene-optimized and fit one representation per posed scene. Agricultural applications demonstrate reconstruction and phenotyping potential \cite{smitt2023pag,jiang2025cotton3dgaussians,zhang2025wheat3dgs}, but appearance alone does not establish photogrammetry-referenced geometry or agronomic utility.

\noindent\textbf{Feed-forward visual geometry.}
Learning-based MVS networks predict depth from aggregated multi-view evidence \cite{wei2021aa} and are trained on large multi-view datasets \cite{yao2020blendedmvs}. Recent feed-forward models instead regress geometry directly: DUSt3R predicts unconstrained point maps \cite{wang2024dust3r}; MASt3R adds grounded matching \cite{leroy2024grounding}; VGGT jointly predicts cameras and geometry \cite{wang2025vggt}; $\pi^3$ (written Pi3 hereafter) targets permutation-equivariant reconstruction \cite{wang2025pi}; and MapAnything predicts metric-scale geometry directly \cite{keetha2025mapanything}. Whether their aligned geometry, which is often accurate, also yields usable metric scale on field crops remains unmeasured.

\noindent\textbf{Benchmark scope and distinction.}
Table~\ref{tab:related_dataset_comparison} contrasts general 3D benchmarks with crop-phenotyping resources. General benchmarks support MVS or novel-view synthesis (NVS) evaluation but cover neither repeated crop development nor agronomic measurements; crop resources offer temporal or multi-view labels but no field-scale 3D benchmark. UAV3DCrop combines repeated multi-directional UAV surveys with appearance, geometry, metric-scale, and canopy-height evaluation.

\begin{table*}[t]
  \caption{Scope of representative general-purpose 3D and crop-phenotyping resources.}
  \label{tab:related_dataset_comparison}
  \centering
  \footnotesize
  \setlength{\tabcolsep}{2.6pt}
  \renewcommand{\arraystretch}{1.03}
  \begin{tabular}{@{}lccclll@{}}
    \toprule
    Dataset & Field UAV & Crops & Temporal coverage & View sampling & 3D benchmark & Agronomic information \\
    \midrule
    ETH3D \cite{schops2017multi} & No & -- & 1 date & DSLR/stereo & MVS vs.\ laser scan & -- \\
    BlendedMVS \cite{yao2020blendedmvs} & No & -- & 1 date & Multi-view & MVS vs.\ rendered depth & -- \\
    Mip-NeRF 360 \cite{barron2022mip} & No & -- & 1 date & 360\textdegree{} trajectory & NVS & -- \\
    GroMo25 \cite{bansal2025gromo25} & No & 4 & Multiple dates & 24$\times$5 views & -- & Age, leaf count \\
    TomatoMAP \cite{zhang2025tomato} & No & 1 & 32 dates & 12$\times$4 views & -- & Growth stage, boxes, masks \\
    MIPDB \cite{wang2024mipdb} & Mixed & 1 & Multiple dates & Ground+aerial & -- & Point-line labels \\
    UAV3DCrop & Yes & 4 & 39 dates, 3 seasons & Nadir+oblique & NVS/depth/pose/points & Canopy height, effective LAI \\
    \bottomrule
  \end{tabular}
\end{table*}

\section{Dataset and Benchmark Protocol}
\label{sec:dataset}

\subsection{Dataset Overview and Acquisition}
\label{sub:dataset_overview}

UAV3DCrop is a public, multi-year, multi-crop, multi-angle UAV RGB dataset collected in production fields in the US Midwest from 2023 to 2025 (Fig. \ref{fig:fig1}). The benchmark spans 91 crop--date--plot scenes observed on 39 dates across eight longitudinal sequences, each covering one crop--year--plot combination (Table~\ref{tab:dataset_inventory}). A scene is one independently flown survey; repeated scenes record seasonal development and are reconstructed separately.

\begin{table}[t]
  \centering
  \small
  \setlength{\tabcolsep}{1.5pt}
  \caption{Dataset inventory by year and crop.}
  \label{tab:dataset_inventory}
  \begin{tabular}{@{}>{\raggedright\arraybackslash}p{0.20\linewidth}rrrrrr@{}}
    \toprule
    & \multicolumn{4}{c}{UAV data} & \multicolumn{2}{c}{Field data} \\
    \cmidrule(lr){2-5}\cmidrule(l){6-7}
    Year and crop & Scenes & RGB & Poses & Depth & Height & LAI \\
    \midrule
    2023 Corn & 11 & 9,406 & 9,396 & 9,406 & -- & 168 \\
    2023 Soybean & 10 & 7,039 & 7,039 & 7,037 & -- & 160 \\
    2024 Corn & 18 & 15,712 & 15,434 & 14,868 & -- & 71 \\
    2025 Corn & 12 & 14,483 & 14,367 & 14,478 & 66 & 66 \\
    2025 Soybean & 12 & 12,997 & 12,997 & 12,997 & 48 & 48 \\
    2025 Wheat & 14 & 19,550 & 19,546 & 19,462 & 60 & 60 \\
    2025 Oat & 14 & 9,643 & 9,634 & 9,642 & 36 & 36 \\
    \midrule
    Total & 91 & 88,830 & 88,413 & 87,890 & 210 & 609 \\
    \bottomrule
  \end{tabular}
\end{table}

Images were acquired with a DJI Mavic 3M using real-time kinematic (RTK) positioning, with nominal horizontal and vertical accuracies of 1 and 1.5 cm. Its RGB camera has a 24~mm-equivalent focal length, an 84\textdegree{} diagonal field of view, and a resolution of $5280 \times 3956$ pixels. Each mission comprised eight oblique flight lines at a gimbal pitch of $-45$\textdegree{} (that is, 45\textdegree{} from nadir) with viewing azimuths spaced 45\textdegree{} apart, plus two mutually perpendicular nadir grids. Oblique and nadir overlap were 70--80\% and approximately 80\%, respectively; flying height was 12.2--18.3 m above ground level.

Ground-based effective LAI, which quantifies foliage density within the canopy, was measured for the benchmark surveys in all three years with an LAI-2200C plant canopy analyzer (LI-COR Biosciences, Lincoln, NE, USA). Measurements were taken on the flight day, or on the nearest available date, under diffuse sky conditions near sunrise or sunset or under overcast skies. Plant height was measured in 2025 only, as the mean of five repeated readings taken at the same sampling point.

\subsection{Pose Processing and Quality Control}
\label{sub:pose_qc}

Each scene was processed independently. The original image metadata provided RTK positions, orientations, and initial camera intrinsics. We initialized geolocation with the RTK references and refined camera parameters through image alignment and bundle adjustment in Agisoft Metashape. Interior and exterior orientation parameters were exported in a nerfstudio-compatible representation \cite{tancik2023nerfstudio}; the dense MVS depth maps generated after SfM and bundle adjustment provide the common photogrammetric reference for $z$-depth evaluation.

Across the 91 benchmark scenes, camera registration ranged from 96.9\% to 100.0\%, average ground sampling distance (GSD) from 3.58 to 5.84 mm px$^{-1}$, and root-mean-square (RMS) reprojection error from 0.93 to 1.98 px (median 1.37 px). The median total camera-location residual, computed between RTK-recorded and bundle-adjusted camera centers, was 2.34 cm; because no independent ground check points were surveyed, this is a measure of internal consistency rather than of external accuracy. Supplementary Sec.~\ref{sup:scene_inventory} provides the complete scene-level QC audit.

\noindent\textbf{Data availability.} The RGB imagery and photogrammetric depth reference are public under CC BY 4.0 without an access request. Because raw camera metadata encode RTK acquisition locations, released camera records retain only relative poses and intrinsics.

\subsection{Two-Track Benchmark}
\label{sub:benchmark_protocols}

\noindent\textbf{Track A: scene-optimized core benchmark.}
We evaluate two NeRF baselines, Nerfacto and Instant-NGP \cite{muller2022instant,tancik2023nerfstudio}, together with five 3DGS baselines: the Nerfstudio Splatfacto and Splatfacto-big configurations, which implement and extend 3D Gaussian splatting \cite{kerbl20233d,tancik2023nerfstudio,nerfstudio_splatfacto_docs}, Mip-Splatting \cite{yu2024mipsplatting}, Scaffold-GS \cite{lu2024scaffoldgs}, and CityGaussian \cite{liu2024citygaussian}. Splatfacto-big tests a larger Gaussian budget.

Methods share a fixed split in every scene: evenly spaced images form a deterministic 10\% test set, and the remaining 90\% are used for optimization. This evaluates view interpolation under dense view sampling, rather than extrapolation beyond the acquired viewing geometry. Unmodified NVS renderings are scored by peak signal-to-noise ratio (PSNR), structural similarity index measure (SSIM), and learned perceptual image patch similarity (LPIPS); throughput is reported in frames per second (FPS).

Photogrammetry-referenced geometry is evaluated as camera-frame $z$-depth in meters, that is, distance along the optical axis rather than along the viewing ray, on the same held-out image raster used for NVS. Metrics are root-mean-square error (RMSE), absolute relative error (AbsRel), scale-invariant logarithmic error (SILog), and Pearson correlation ($r$).

Canopy-height recovery provides downstream agronomic validation for methods that output explicit geometry. The matched subset contains 31 scenes and 210 field sampling points. Within a 0.4~m horizontal radius of each sampling point, canopy height is the difference between the canopy-surface height and the local ground height, that is, a local canopy height model. Ground height is the 50th percentile of points from a separate bare-ground survey of the same plot, reconstructed by the same method; canopy height is the 85th percentile of crop-date points for corn, soybean, and wheat and the 90th for oat. Both point sets require at least 20 points, and every method supplies its own bare-ground reference, so no method depends on another's reconstruction. Because the field reference averages five readings at one sampling point, it likewise characterizes canopy height over that neighborhood, and the two are treated as comparable at the plot scale sampled here. Predictions are scored using RMSE, mean absolute error (MAE), and the coefficient of determination ($R^2$). Errors at sampling points are macro-averaged by scene; $R^2$ uses all 210 pairs. Supplementary Sec.~\ref{sup:height_validation} reports additional analyses.

Both geometric tasks are evaluated twice: once with the native depth export of each method and once with a revised export, using the same $z$-depth definition throughout. Reusing the same checkpoints and training runs, the revised export discards NeRF samples falling outside an axis-aligned scene bounding box (AABB) scaled by $1.25\times$, and masks Gaussians whose centers lie outside that box or whose longest physical axis exceeds 2 m. We selected these two thresholds in preliminary output-control tests and then fixed them across all scenes and methods. The revised export also feeds canopy-height recovery, whereas NVS always uses unmodified RGB renderings. Per-method mean valid-pixel coverage is 98.604--99.995\%; a single method--scene result falls below 95\%. Supplementary Sec.~\ref{sup:depth_revision} gives per-method audits and native-to-revised results.

\noindent\textbf{Track B: zero-shot feed-forward evaluation.}\par
\noindent We evaluate MASt3R \cite{leroy2024grounding}, VGGT \cite{wang2025vggt}, Pi3 \cite{wang2025pi}, and MapAnything \cite{keetha2025mapanything} using official pretrained weights without crop-specific fine-tuning. From each scene we draw 140 random subsets of 36 images each, downsample every image by a factor of eight per axis, and process each subset independently; the 140 subset scores are then averaged into one scene result. Model-specific heads recover cameras, per-view $z$-depth, point maps, and ray directions. MapAnything also predicts an explicit metric scale.

Models receive only the RGB subsets; reference poses, sparse points, and dense geometry are reserved for evaluation. Predicted trajectories are aligned to the RTK/SfM reference by a closed-form similarity (Umeyama) fit before we compute the root-mean-square absolute trajectory error (ATE RMSE) and the pose-accuracy area under the curve at a 5\textdegree{} threshold (AUC@5). Point maps and $z$-depth are scored by AbsRel and by the inlier rate under a $\delta<1.03$ threshold, where $\delta$ is the larger of the prediction-to-reference and reference-to-prediction ratios. Geometry is scored both after a per-scene least-squares scale-and-shift alignment and at the model's unaligned metric scale, separating structural accuracy from metric-scale recovery.

\noindent\textbf{Reporting and reuse protocol.}
The tracks address complementary questions and are therefore reported separately. Within Track A, metrics and canopy-height errors are first aggregated by scene to give each survey equal weight. Within Track B, subset results are averaged by scene and then by sequence to give each sequence equal weight.

\noindent\textbf{Uncertainty.}
We compare each numerical leader with the runner-up using 20,000 paired bootstrap replicates and 95\% percentile intervals. NVS and depth resample the 91 paired scenes. Height resamples the 31 paired scenes while retaining sampling points within scene, and feed-forward evaluation resamples paired scenes within each of the eight sequences before recomputing the equal-sequence average. A leader is reported as statistically supported when the interval excludes zero; otherwise, the two methods are reported as tied. Because the bootstrap resamples scenes within sequences, it distinguishes a stable leader from a gap that reflects only which scenes happened to be sampled.

\noindent\textbf{Scene-condition diagnostics.}
Each scene and feed-forward subset is reconstructed independently, and repeated acquisitions index scene conditions across the eight sequences. We analyze four degradation-oriented outcomes---negative NVS PSNR, depth RMSE, negative pose AUC@5, and point-map AbsRel---all oriented so that larger values are worse. The stressors are days since first acquisition; negative log image count, $-\log(1+n_{\mathrm{images}})$; GSD; negative tie-point multiplicity, the mean number of images observing each triangulated tie point; and RMS reprojection error. Feed-forward subset estimates are first averaged by scene.

For each method--outcome--stressor combination, we fit an ordinary least-squares model to standardized response and predictor values with sequence fixed effects. Models for image count, GSD, tie-point multiplicity, and reprojection error also include days since first acquisition to account for sequence progression. We calculate acquisition-date-clustered standard errors over 39 dates and two-sided $p$ values from $t$ statistics with 38 degrees of freedom. Benjamini--Hochberg correction is performed separately for each outcome family: 35 method--stressor tests for NVS and depth ($7\times5$) and 20 tests for pose and point-map geometry ($4\times5$). For ordered raw values $p_{(1)}\le\cdots\le p_{(m)}$, the adjusted values ($q$ values) are $q_{(i)}=\min_{j\ge i}\{mp_{(j)}/j,1\}$, and coefficients with $q<0.05$ are reported as supported at a false discovery rate (FDR) of 5\%. Each $\beta$ is a standardized regression coefficient, so positive $\beta$ denotes degradation under greater measured stress. These coefficients are a diagnostic association screen, not a causal analysis.

\section{Results}
\label{sec:results}

Across ranked tables, bold marks the numerical best and underlining marks the runner-up; rankings use unrounded values. In the main tables, light-blue shading marks a leader whose best-versus-runner-up paired 95\% bootstrap interval excludes zero. Full intervals and win rates are reported in Supplementary Sec.~\ref{sup:rank_uncertainty}.

\subsection{Scene-optimized reconstruction (RQ1)}
\label{sub:scene_optimized_results}

\paragraph{Novel-view synthesis.}
\label{sub:NVS_section}

\begin{figure*}[t]
  \centering
  \includegraphics[width=0.80\textwidth]{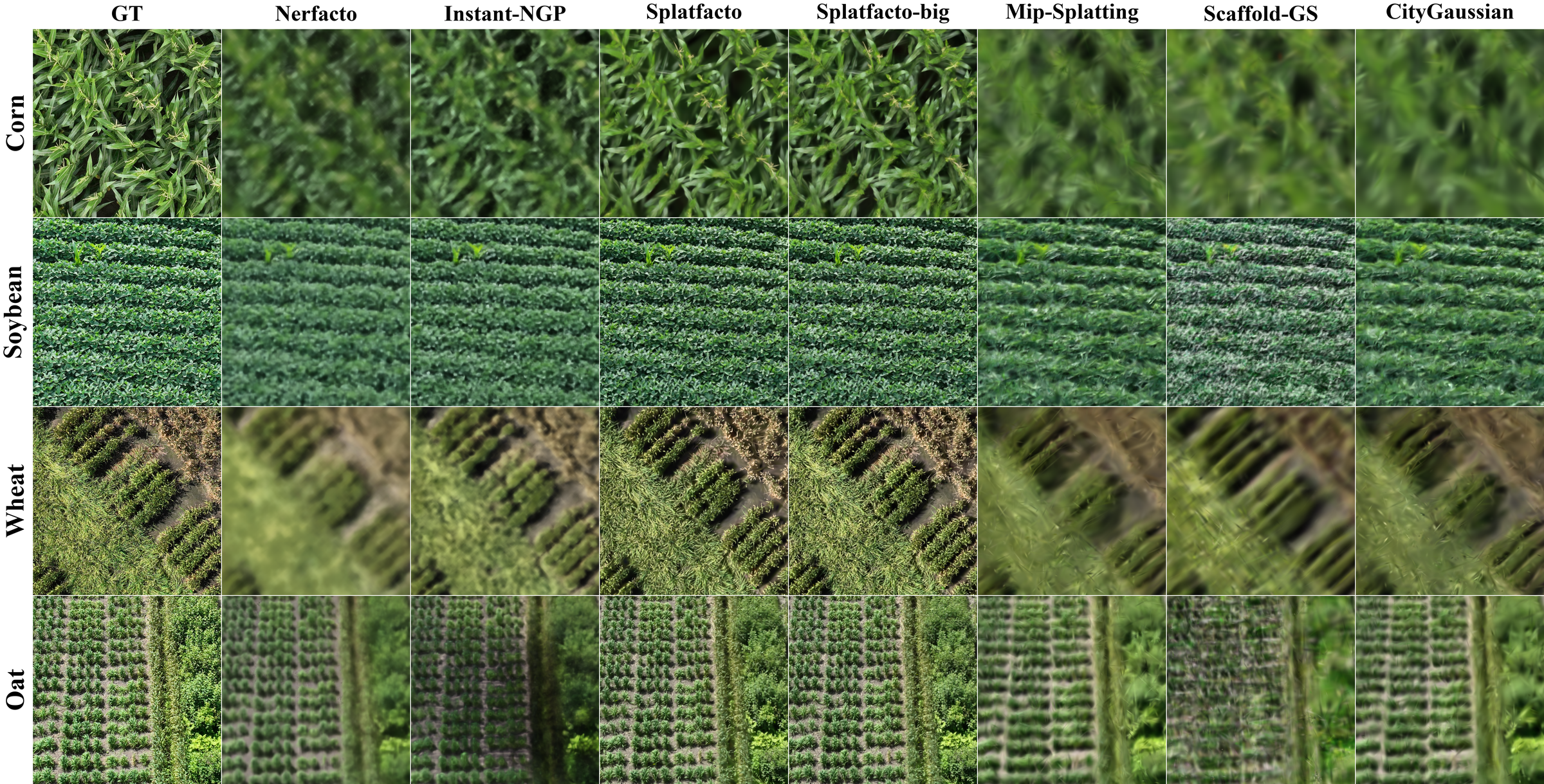}
  \caption{Qualitative NVS comparison on held-out views, using matched image regions. Rows: crops; columns: reference and seven scene-optimized methods.}
  \Description{A four-by-eight image grid compares ground-truth crop-field regions with Nerfacto, Instant-NGP, Splatfacto, Splatfacto-big, Mip-Splatting, Scaffold-GS, and CityGaussian for corn, soybean, wheat, and oat.}
  \label{fig:nvs_qualitative}
\end{figure*}

Splatfacto and Splatfacto-big most consistently preserve narrow leaves, canopy boundaries, and repeated rows (Fig.~\ref{fig:nvs_qualitative}). The NeRF baselines recover broad canopy layout but smooth thin leaves and local texture, and the remaining Gaussian variants hold coarse structure with greater blur, smearing, or clutter.

Splatfacto-big leads every appearance metric at 19.40 dB PSNR, with Splatfacto second but fastest at 25.15 FPS (Table~\ref{tab:nvs_scene_optimized}). The remaining methods fall to 15.28--16.42 dB, a gap of about 3 dB. The larger Gaussian budget adds 0.35 dB over Splatfacto on unrounded values but reduces throughput by 57\%.

Crop-specific PSNR preserves the aggregate ordering (Supplementary Table~\ref{tab:sup_tracka_crop}). Splatfacto-big ranks first and Splatfacto second on all four crops, with gaps of 0.27--0.49 dB. Corn has the lowest PSNR for six of the seven methods; for Instant-NGP the lowest crop is oat.

\begin{table}[t]
  \caption{Scene-macro NVS results from native RGB renderings.}
  \label{tab:nvs_scene_optimized}
  \centering
  \small
  \setlength{\tabcolsep}{3pt}
  \begin{tabular}{@{}lrrrr@{}}
    \toprule
    Method & PSNR (dB) $\uparrow$ & SSIM $\uparrow$ & LPIPS $\downarrow$ & FPS $\uparrow$\\
    \midrule
    Nerfacto             & 15.72 & 0.185 & 0.923 & 0.04 \\
    Instant-NGP          & 15.99 & 0.206 & 0.836 & 0.01 \\
    Splatfacto           & \tabsecond{19.04} & \tabsecond{0.495} & \tabsecond{0.403} & \tabsupported{25.15} \\
    Splatfacto-big       & \tabsupported{19.40} & \tabsupported{0.546} & \tabsupported{0.318} & 10.82 \\
    Mip-Splatting        & 16.42 & 0.231 & 0.677 & \tabsecond{23.74} \\
    Scaffold-GS          & 15.28 & 0.213 & 0.694 & 17.22 \\
    CityGaussian         & 16.40 & 0.229 & 0.872 & 8.94 \\
    \bottomrule
  \end{tabular}
\end{table}

\paragraph{Depth reconstruction.}
\label{sub:depth_section}

Depth reverses this ordering (Table~\ref{tab:depth_scene_optimized}). Scaffold-GS leads all four metrics, reaching 0.722 m RMSE against 0.934--1.544 m for the rest, with CityGaussian second and every leader-versus-runner-up interval excluding zero.

Scaffold-GS also leads on every crop, from 0.513 m on wheat to 0.909 m on corn, with CityGaussian second throughout (Supplementary Table~\ref{tab:sup_tracka_crop}). Corn is the hardest crop for every method.

\begin{table}[t]
  \caption{Scene-macro $z$-depth reconstruction under the revised export.}
  \label{tab:depth_scene_optimized}
  \centering
  \small
  \setlength{\tabcolsep}{3pt}
  \begin{tabular}{@{}lrrrr@{}}
    \toprule
    Method & RMSE (m) $\downarrow$ & AbsRel $\downarrow$ & SILog $\downarrow$ & Pearson $r$ $\uparrow$ \\
    \midrule
    Nerfacto             & 1.117 & 0.035 & 0.040 & 0.873 \\
    Instant-NGP          & 1.525 & 0.037 & 0.061 & 0.836 \\
    Splatfacto           & 1.379 & 0.042 & 0.058 & 0.831 \\
    Splatfacto-big       & 1.544 & 0.047 & 0.066 & 0.820 \\
    Mip-Splatting        & 0.934 & 0.035 & 0.031 & 0.895 \\
    Scaffold-GS          & \tabsupported{0.722} & \tabsupported{0.025} & \tabsupported{0.025} & \tabsupported{0.906} \\
    CityGaussian         & \tabsecond{0.849} & \tabsecond{0.032} & \tabsecond{0.029} & \tabsecond{0.898} \\
    \bottomrule
  \end{tabular}
\end{table}

\paragraph{Appearance--geometry relationship.}
\label{sub:cross_task_results}

\begin{figure}[b]
  \centering
  \includegraphics[width=0.55\textwidth]{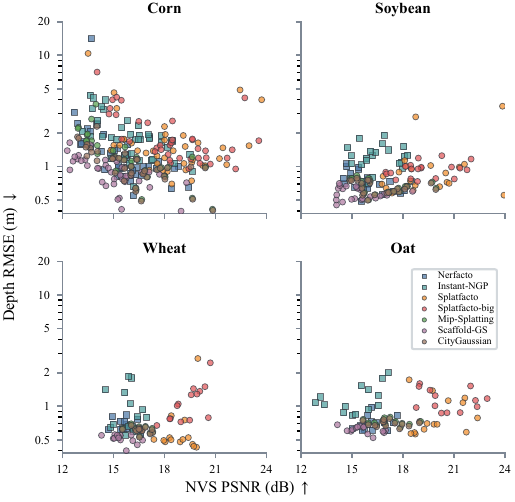}
  \caption{Scene-level native NVS PSNR versus revised-export depth RMSE, faceted by crop. Each marker is one method--scene pair; squares denote NeRF and circles 3DGS methods. Depth-RMSE axes are logarithmic; rightward and downward are better.}
  \Description{Four crop-specific scatter plots compare scene-level NVS PSNR and depth RMSE for seven reconstruction methods. Colors identify methods, squares identify NeRF methods, and circles identify 3DGS methods.}
  \label{fig:cross_task_map}
\end{figure}

Splatfacto-big has the highest scene-mean PSNR on every crop, whereas Scaffold-GS has the lowest depth RMSE (Fig.~\ref{fig:cross_task_map}). Increasing the Splatfacto Gaussian budget raises PSNR throughout but worsens depth on corn, wheat, and oat, with only a marginal improvement on soybean. Higher appearance quality therefore does not imply lower geometric error.

\paragraph{Sensitivity to the revised export.}
The revised export helps most where native depth was worst: Splatfacto improves from 7.498 to 1.379 m RMSE, with smaller gains for Splatfacto-big and Instant-NGP, whereas Mip-Splatting and CityGaussian move by under 0.17 m (Supplementary Table~\ref{tab:sup_depth_configs}). It therefore removes large depth outliers without changing training or NVS.

\subsection{Canopy-height validation (RQ2)}
\label{sub:PH_section}

Scaffold-GS and Splatfacto are effectively tied for canopy height, at 0.091 and 0.092 m scene-macro MAE, and all three paired intervals include zero (Table~\ref{tab:plant_height_scene_balanced}). Nerfacto follows, while the remaining methods reach only 0.156--0.197 m.

\begin{table}[t]
  \caption{Canopy-height estimation against the field plant-height reference. MAE and RMSE are scene-macro means; $R^2$ uses all paired predictions.}
  \label{tab:plant_height_scene_balanced}
  \centering
  \small
  \setlength{\tabcolsep}{4pt}
  \begin{tabular}{@{}lrrr@{}}
    \toprule
    Method & MAE (m) $\downarrow$ & RMSE (m) $\downarrow$ & $R^2$ $\uparrow$ \\
    \midrule
    Nerfacto             & 0.136 & 0.151 & 0.960 \\
    Instant-NGP          & 0.197 & 0.214 & 0.897 \\
    Splatfacto           & \tabsecond{0.092} & \tabsecond{0.099} & \tabsecond{0.978} \\
    Splatfacto-big       & 0.156 & 0.219 & 0.863 \\
    Mip-Splatting        & 0.165 & 0.176 & 0.940 \\
    Scaffold-GS          & \tabbest{0.091} & \tabbest{0.099} & \tabbest{0.980} \\
    CityGaussian         & 0.172 & 0.182 & 0.937 \\
    \bottomrule
  \end{tabular}
\end{table}

The crop-level height ranking differs from the crop-level depth ranking (Supplementary Table~\ref{tab:sup_height_crop}). Oat has the highest canopy-height MAE for every method, spanning 0.177--0.458 m. Scaffold-GS leads wheat and oat, whereas Splatfacto leads corn and soybean. No single method is best on every crop, so pooled height scores conceal crop-specific behavior.

The effect of the revised export on canopy-height error also varies by method (Supplementary Table~\ref{tab:sup_height_configs}). Splatfacto gains most, with pooled MAE decreasing from 0.350 to 0.089 m. Splatfacto-big and CityGaussian also improve, whereas Nerfacto, Scaffold-GS, and Instant-NGP change little. All methods retain finite predictions at all 210 sampling points.

Scaffold-GS and Splatfacto retain date-demeaned $R^2$ values of 0.972 and 0.969, but within-scene correlations are 0.496 and 0.525. Broad height differences are therefore recovered more reliably than fine within-scene ordering (Supplementary Sec.~\ref{sup:height_validation}). Scaffold-GS leads both depth and height, but the ranking below it reorders, showing that depth accuracy does not fully determine downstream utility.

\subsection{Zero-shot feed-forward evaluation (RQ3)}
\label{sub:feedforward_results}

MapAnything leads on seven of the eight metrics, while Pi3 has the lowest ray-direction error; every top-versus-runner-up interval excludes zero (Table~\ref{tab:feedforward_zero_shot}). The largest separation is absolute scale: MapAnything obtains 0.027 AbsRel, whereas the other models reach 0.890--0.965 despite far more accurate aligned geometry. This comparison should be interpreted in light of model design: only MapAnything includes a dedicated metric-scale head, whereas the other models predict normalized geometry. Alignment can therefore conceal severe metric-scale failure.

\begin{figure}[t]
  \centering
  \includegraphics[width=0.55\textwidth]{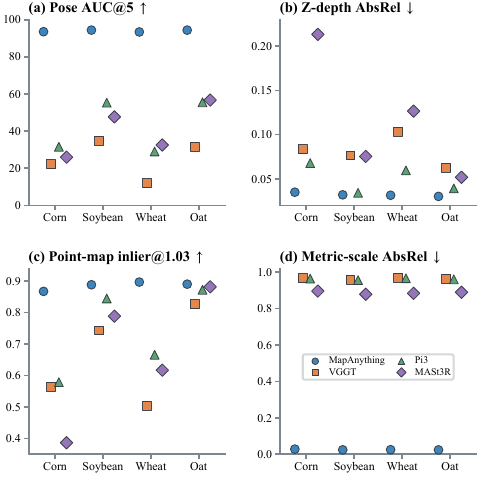}
  \caption{Zero-shot pose, depth, point-map, and metric-scale accuracy grouped by crop. Markers average four corn, two soybean, one wheat, and one oat sequence.}
  \Description{Four dot plots compare pose, depth, point-map, and metric-scale accuracy for MapAnything, VGGT, Pi3, and MASt3R after grouping the eight sequences by crop.}
  \label{fig:feedforward_crops}
\end{figure}

MapAnything barely varies across crops, holding pose AUC@5 within one percentage point, whereas MASt3R swings fourfold in $z$-depth AbsRel between oat and corn, and VGGT and Pi3 lose pose and point-map accuracy on wheat (Fig.~\ref{fig:feedforward_crops}; Supplementary Table~\ref{tab:sup_feedforward_crop}).

\begin{table*}[!t]
  \caption{Zero-shot camera and geometry estimation. Scene metrics are averaged within sequence and then macro-averaged over eight sequences.}
  \label{tab:feedforward_zero_shot}
  \centering
  \small
  \setlength{\tabcolsep}{3.2pt}
  \begin{tabular}{@{}lrrrrrrrr@{}}
    \toprule
    & Scale & \multicolumn{2}{c}{Point map} & \multicolumn{2}{c}{Pose} & \multicolumn{2}{c}{$z$-depth} & Ray \\
    \cmidrule(lr){2-2}\cmidrule(lr){3-4}\cmidrule(lr){5-6}\cmidrule(lr){7-8}\cmidrule(l){9-9}
    Method & AbsRel $\downarrow$ & AbsRel $\downarrow$ & Inlier@1.03 $\uparrow$ & ATE RMSE $\downarrow$ & AUC@5 (\%) $\uparrow$ & AbsRel $\downarrow$ & Inlier@1.03 $\uparrow$ & Error (\textdegree{}) $\downarrow$ \\
    \midrule
    MapAnything & \tabsupported{0.027} & \tabsupported{0.037} & \tabsupported{0.879} & \tabsupported{0.006} & \tabsupported{93.8} & \tabsupported{0.033} & \tabsupported{0.415} & 2.28 \\
    VGGT        & 0.965 & 0.099 & 0.634 & 0.086 & 25.2 & 0.082 & 0.263 & \tabsecond{1.64} \\
    Pi3         & 0.963 & \tabsecond{0.084} & \tabsecond{0.693} & \tabsecond{0.051} & \tabsecond{40.2} & \tabsecond{0.055} & \tabsecond{0.389} & \tabsupported{1.42} \\
    MASt3R      & \tabsecond{0.890} & 0.207 & 0.578 & 0.164 & 36.1 & 0.148 & 0.323 & 3.95 \\
    \bottomrule
  \end{tabular}
\end{table*}

\subsection{Sensitivity to temporal and scene conditions}
\label{sub:temporal_variation_section}

\begin{figure*}[!t]
  \centering
  \includegraphics[width=\textwidth]{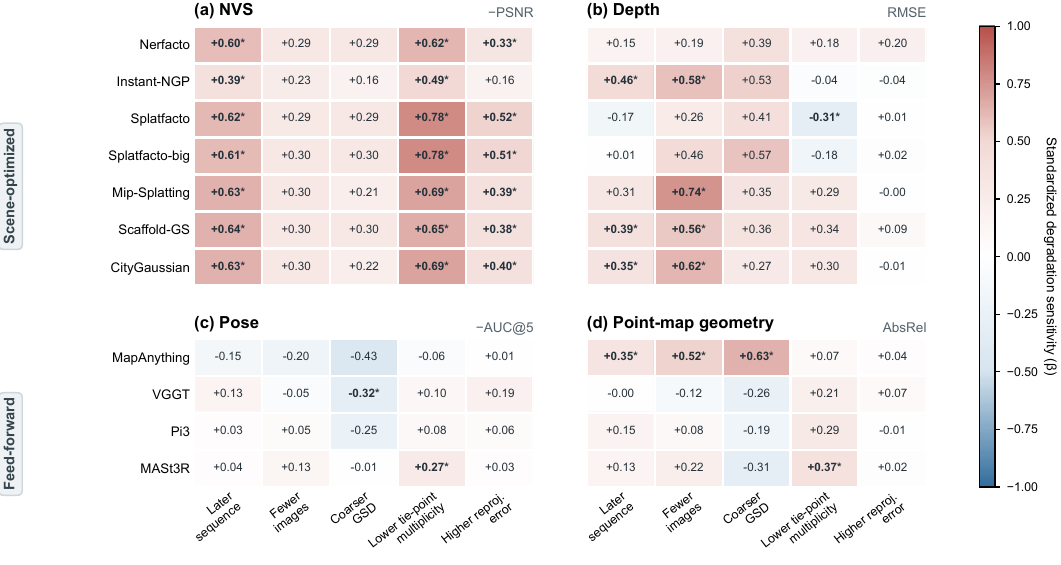}
  \caption{Method-specific sensitivity to temporal and measured scene conditions. Each cell is a standardized coefficient $\beta$ from a separate 91-scene model; stressors and outcomes both point toward degradation, so positive $\beta$ is worse. All fits include sequence fixed effects, and all but the sequence-position models also control for sequence progression. Asterisks mark Benjamini--Hochberg FDR $q<0.05$ within each outcome family (35 tests for NVS and depth; 20 for pose and point maps), using acquisition-date-clustered standard errors.}
  \Description{Four heatmaps compare standardized degradation sensitivities for seven scene-optimized and four feed-forward methods over 91 scenes per method. The upper panels show NVS and depth; the lower panels show pose and point-map geometry. Red cells indicate greater degradation under greater stress, blue cells indicate the opposite association, and asterisks mark FDR-supported coefficients.}
  \label{fig:scene_condition_sensitivity}
\end{figure*}

Acquisition and SfM conditions are associated with degradation to different degrees across methods and outputs (Fig.~\ref{fig:scene_condition_sensitivity}).

NVS has the most consistent failure profile: all 35 coefficients are positive. Later sequence position and lower tie-point multiplicity retain FDR support for every method, and higher reprojection error for six. Their median $\beta$ values are 0.616, 0.689, and 0.394; no image-count or GSD coefficient reaches FDR support ($q\ge0.05$). Sequence progression and tie-point multiplicity therefore identify a shared appearance-failure regime.

Depth is more method-dependent. Fewer images produce the strongest median sensitivity ($\beta=0.562$), with FDR support for four methods; later sequence position is supported for three. Coarser GSD is positive throughout but does not survive FDR correction, while reprojection error and tie-point multiplicity are mixed. Depth degradation is therefore driven primarily by limited image count.

Evidence for feed-forward pose sensitivity is weak and inconsistent: only 2 of 20 cells retain FDR support. Point-map geometry retains four supported cells, each specific to one model. MapAnything degrades at later positions, with fewer images, and under coarser GSD; MASt3R is sensitive to lower tie-point multiplicity. The supported stressors differ across models, reinforcing output- and method-specific failure modes.

\section{Discussion and Conclusion}
\label{sec:discussion}

The central result is a task-conditional method ordering. Splatfacto-big leads appearance on every crop, whereas Scaffold-GS leads depth throughout and is numerically strongest for canopy height, where the appearance leader's error is 71\% higher.

This disagreement follows from what the tasks measure. Held-out NVS rewards image formation at observed viewpoints; $z$-depth tests the recovered surface against a common photogrammetric reference; and canopy height is a local canopy-to-ground difference. A representation can therefore reproduce color and texture while placing geometry incorrectly, and conversely, shared vertical offsets can cancel in a height difference even when depth error remains. Appearance-focused applications therefore call for a different operating point than metric mapping or phenotyping does.

Crop-grouped results reveal task-specific failure regimes. Corn has the highest scene-optimized depth error, oat is hardest for canopy-height recovery, and feed-forward failures depend on the model and output, which argues for crop-grouped reporting alongside pooled scores.

Method-specific sensitivities further separate the targets: NVS degrades at later sequence positions and with lower tie-point multiplicity, depth is most often sensitive to fewer images, and feed-forward responses are model-specific. Image count and SfM diagnostics therefore provide practical cues for identifying acquisitions that merit inspection.

The feed-forward results are encouraging but reveal substantial domain shift. MapAnything is the only model that is both stable across crops and metrically calibrated, whereas VGGT, Pi3, and MASt3R vary more and fail to recover absolute scale. Alignment conceals this failure: an aligned point map may preserve shape while being unusable for measurements in meters, so both aligned and unaligned metric-scale outputs must be reported when agricultural use depends on physical dimensions. Zero-shot inference enables rapid transfer testing, whereas scene optimization remains the stronger option when metric accuracy is required.

\subsection{Limitations and future work}
Three design choices bound the interpretation of these results. First, SfM-derived poses and photogrammetry-referenced geometry are least reliable in repetitive, textureless, or moving scenes \cite{iglhaut2019structure}; independent laser-scan validation would strengthen future benchmarks. Second, the current NVS split evaluates only view interpolation, motivating future held-out-azimuth and sparse-view splits. Third, field measurements are uneven across years and were not recorded on a standard growth-stage scale: plant height is available only for 2025, and because effective LAI rises with acquisition date, its apparent association with scene quality cannot be separated from sequence progression (Supplementary Fig.~\ref{fig:sup_lai_scene_quality}).

The two tracks motivate distinct priorities. Scene-optimized methods could combine photometric consistency with metric-depth and canopy-surface regularization, and temporal priors should then be tested to determine whether they improve repeated reconstructions without suppressing genuine growth. Feed-forward models require controlled variation in view count and azimuthal coverage, followed by adaptation tests across crops, sites, and years. Linked effective-LAI measurements open a further target: every metric reported here probes the canopy surface, whereas LAI summarizes foliage density within the canopy, so retrieving LAI from a learned representation \cite{yang2025nerflai} would test whether a visually plausible scene also reproduces canopy interior structure.

\subsection{Benchmark reuse and broader impacts}
UAV3DCrop enables research on faster reconstruction, crop adaptation, and temporal scene modeling. We recommend that reuse preserve the fixed manifests and scene-level grouping, report native and revised-export Track A geometry separately, retain both aligned and unaligned metric-scale Track B results, and disclose any crop-specific fine-tuning. For future learned models, complete sequences should be assigned to either training or testing to prevent leakage between adjacent surveys.

Looking ahead, field-scale 3D reconstruction could extend tasks that 2D products address only indirectly. High-throughput phenotyping can support breeding selection that still relies partly on manual assessment \cite{chivasa2020uav,jin2020high}; reconstructed canopy architecture may provide a complementary structural trait. Organ-scale 3D traits can sharpen disease assessment \cite{yang20243d}. Likewise, because multi-temporal remote sensing already informs within-season management \cite{mulla2013twenty}, repeated geometry could test whether 3D growth rates add value beyond a single observation date. Climate change increases the need for resilient agricultural monitoring and management \cite{azadi2021rethinking}, while the monitoring implication itself remains a prospective application of the benchmark. Overall, dependable crop-field reconstruction requires joint evaluation of appearance, geometry, canopy height, and metric scale.

\section*{Generative AI Usage}
OpenAI ChatGPT and Codex helped debug analysis code, prepare figures, and edit the language of this manuscript. The authors reviewed all such output and are solely responsible for the content.

\clearpage
\bibliographystyle{unsrtnat}
\bibliography{main}

\clearpage
\appendix
\section{Supplementary Material}
\label{sec:suppl}

\subsection{Scene-optimized results grouped by crop}
\label{sup:crop_results}

Table~\ref{tab:sup_tracka_crop} reports the Track A metrics after grouping scenes by crop.

\begin{table*}[!t]
  \caption{Track A results grouped by crop. NVS uses native RGB renderings and depth uses the revised geometry output. Each entry is the equal-weight mean of scene metrics within that crop. Scene counts are corn 41, soybean 22, wheat 14, and oat 14.}
  \label{tab:sup_tracka_crop}
  \centering
  \small
  \setlength{\tabcolsep}{3.5pt}
  \begin{tabular}{@{}lrrrrrrrr@{}}
    \toprule
    & \multicolumn{4}{c}{NVS PSNR (dB) $\uparrow$} & \multicolumn{4}{c}{Depth RMSE (m) $\downarrow$} \\
    \cmidrule(lr){2-5}\cmidrule(l){6-9}
    Method & Corn & Soybean & Wheat & Oat & Corn & Soybean & Wheat & Oat \\
    \midrule
    Nerfacto       & 15.41 & 15.98 & 15.69 & 16.25 & 1.608 & 0.736 & 0.681 & 0.710 \\
    Instant-NGP    & 16.06 & 16.45 & 15.85 & 15.22 & 1.970 & 1.244 & 0.986 & 1.201 \\
    Splatfacto     & \tabsecond{18.61} & \tabsecond{19.34} & \tabsecond{18.88} & \tabsecond{20.03} & 1.983 & 1.014 & 0.685 & 0.877 \\
    Splatfacto-big & \tabbest{18.88} & \tabbest{19.69} & \tabbest{19.37} & \tabbest{20.49} & 2.114 & 1.001 & 1.133 & 1.135 \\
    Mip-Splatting  & 15.98 & 16.68 & 16.59 & 17.14 & 1.251 & 0.674 & 0.622 & 0.725 \\
    Scaffold-GS    & 15.05 & 15.34 & 15.51 & 15.64 & \tabbest{0.909} & \tabbest{0.573} & \tabbest{0.513} & \tabbest{0.616} \\
    CityGaussian   & 15.96 & 16.64 & 16.59 & 17.12 & \tabsecond{1.086} & \tabsecond{0.655} & \tabsecond{0.610} & \tabsecond{0.700} \\
    \bottomrule
  \end{tabular}
\end{table*}

The same NVS and depth leaders are observed for each crop, but the error magnitudes vary: corn has the highest depth RMSE for every method. Splatfacto-big improves PSNR over Splatfacto in all four groups, whereas its depth change is adverse in three and only marginally favorable on soybean.

\subsection{Output configurations and sensitivity}
\label{sup:depth_revision}

\noindent\textbf{Shared setting.}
The native and revised exports apply only to geometry outputs. The native export applies no geometric validity control; the revised export adds the controls below during depth generation and is also used for the revised canopy-height estimate. The 1.25$\times$ AABB expansion and 2 m Gaussian-axis cutoff were selected in preliminary output-control tests and then frozen for all 91 scenes; neither is retuned by scene or method. All NVS tables and figures use unmodified RGB renderings, with no geometry post-processing.

\noindent\textbf{Common depth definition.}
All methods are evaluated as camera-frame $z$-depth in meters using the same cameras, held-out frames, image raster, and metric implementation. Zero and non-finite predictions are invalid, and no optional minimum/maximum depth clipping is used in the primary table.

\noindent\textbf{Revised NeRF depth controls.}
Native and revised exports use the same checkpoint and depth-output definition. The revised export additionally marks predictions outside the scene AABB, expanded by 1.25, as invalid. The operation is applied only to exported depth and does not change RGB rendering.

\noindent\textbf{Revised Gaussian controls.}
For the Gaussian-based methods, revised depth generation temporarily suppresses a Gaussian when its center lies outside the 1.25$\times$ scene AABB or its largest physical axis exceeds 2 m; opacity is restored after rendering and checkpoints are not edited. This geometry-only post-processing targets floating or oversized Gaussians and is not applied to RGB rendering.

\noindent\textbf{Post-QC coverage audit.}
Prediction-only coverage ranges from 98.604\% to 99.995\% across methods (Table~\ref{tab:sup_depth_coverage}). The sole method--scene result below 95\% is Nerfacto on 2024/Day056\_Corn1 (81.311\%), which remains in all summaries. Final zeros combine QC removals and pre-existing invalid predictions, so the audit reports output completeness after all controls.

\begin{table*}[!t]
  \caption{Prediction-only valid-pixel audit after final depth QC for all seven representative methods on the fixed 91-scene benchmark inventory. Coverage is the number of pixels with retained positive depth divided by the total number of image pixels. Final invalid pixels are stored zeros after QC, so coverage summarizes retained positive depth after all controls.}
  \label{tab:sup_depth_coverage}
  \centering
  \small
  \setlength{\tabcolsep}{7pt}
  \begin{tabular}{@{}lrrrr@{}}
    \toprule
    Method & Retained pixels & Final invalid pixels & Coverage (\%) & Min. scene (\%) \\
    \midrule
    Nerfacto           & 182{,}531{,}724{,}446 & 1{,}405{,}185{,}634 & 99.236 & 81.311 \\
    Instant-NGP        & 181{,}369{,}091{,}971 & 2{,}567{,}818{,}109 & 98.604 & 97.180 \\
    Splatfacto         & 183{,}185{,}460{,}301 &   751{,}449{,}779 & 99.591 & 98.175 \\
    Splatfacto-big     & 183{,}518{,}948{,}756 &   417{,}961{,}324 & 99.773 & 99.305 \\
    Mip-Splatting      & 183{,}927{,}749{,}776 &     9{,}160{,}304 & 99.995 & 99.921 \\
    Scaffold-GS        & 183{,}862{,}658{,}068 &    74{,}252{,}012 & 99.960 & 99.448 \\
    CityGaussian       & 183{,}900{,}580{,}280 &    36{,}329{,}800 & 99.980 & 99.466 \\
    \bottomrule
  \end{tabular}
\end{table*}

Table~\ref{tab:sup_depth_configs} compares native and revised depth exports on the same 91 scenes. Mean RMSE reductions range from 0.159 m for Mip-Splatting to 6.119 m for Splatfacto; the next largest reductions occur for Splatfacto-big and Instant-NGP. Paired bootstrap intervals exclude zero for all seven methods. The revised constraint therefore suppresses depth outliers to a method-dependent degree.

\begin{table*}[!t]
  \caption{Native $\rightarrow$ revised $z$-depth results for the seven representative methods on the 91 benchmark scenes. Lower is better for RMSE, AbsRel, and SILog; higher is better for Pearson $r$.}
  \label{tab:sup_depth_configs}
  \centering
  \small
  \setlength{\tabcolsep}{5pt}
  \begin{tabular}{@{}lcccc@{}}
    \toprule
    Method & RMSE (m) & AbsRel & SILog & Pearson $r$ \\
    \midrule
    Nerfacto           & 2.7193 $\rightarrow$ 1.1168 & 0.0447 $\rightarrow$ 0.0349 & 0.0480 $\rightarrow$ 0.0399 & 0.8318 $\rightarrow$ 0.8728 \\
    Instant-NGP        & 3.4499 $\rightarrow$ 1.5247 & 0.0483 $\rightarrow$ 0.0367 & 0.0860 $\rightarrow$ 0.0612 & 0.7856 $\rightarrow$ 0.8359 \\
    Splatfacto         & 7.4981 $\rightarrow$ 1.3791 & 0.2494 $\rightarrow$ 0.0419 & 0.2392 $\rightarrow$ 0.0584 & 0.5642 $\rightarrow$ 0.8311 \\
    Splatfacto-big     & 4.9068 $\rightarrow$ 1.5436 & 0.1724 $\rightarrow$ 0.0468 & 0.1717 $\rightarrow$ 0.0661 & 0.6758 $\rightarrow$ 0.8204 \\
    Mip-Splatting      & 1.0928 $\rightarrow$ 0.9337 & 0.0391 $\rightarrow$ 0.0349 & 0.0362 $\rightarrow$ 0.0311 & 0.8835 $\rightarrow$ 0.8952 \\
    Scaffold-GS        & 1.3264 $\rightarrow$ 0.7217 & 0.0431 $\rightarrow$ 0.0248 & 0.0562 $\rightarrow$ 0.0254 & 0.7747 $\rightarrow$ 0.9057 \\
    CityGaussian       & 1.0154 $\rightarrow$ 0.8491 & 0.0373 $\rightarrow$ 0.0322 & 0.0319 $\rightarrow$ 0.0285 & 0.8915 $\rightarrow$ 0.8980 \\
    \bottomrule
  \end{tabular}
\end{table*}

\subsection{Canopy-height validation beyond scene means}
\label{sup:height_validation}

The height subset contains 210 measurements in 31 scenes: 60 sampling points in six wheat scenes, 36 in six oat scenes, 66 in 11 corn scenes, and 48 in eight soybean scenes. Table~\ref{tab:sup_height_crop} reports scene-balanced errors separately by crop and Table~\ref{tab:sup_height_levels} complementary pooled metrics. Date-demeaned $R^2$ removes the calendar-date mean, while within-scene Pearson $r$ removes each crop--date--plot mean to test finer spatial ordering.

\begin{figure*}[!t]
  \centering
  \includegraphics[width=0.82\textwidth]{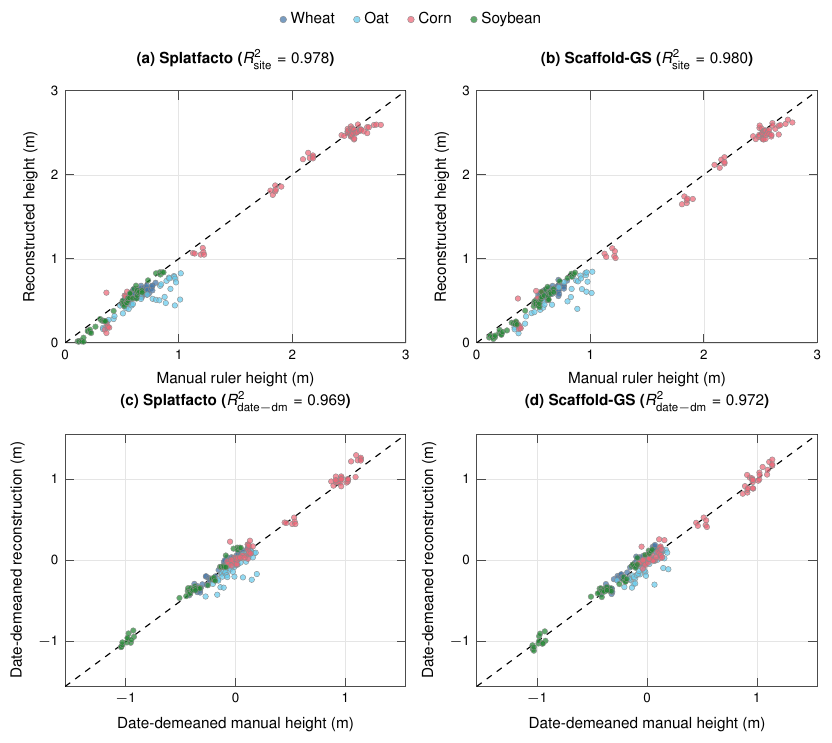}
  \caption{Point-level canopy-height validation for the two numerically leading methods. Panels (a,b) show reconstructed height versus the field plant-height reference for all 210 predictions from each method; panels (c,d) show the same observations after subtracting the corresponding calendar-date mean from both axes. Colors identify crops, and dashed lines denote identity.}
  \Description{Four scatter plots show Splatfacto and Scaffold-GS canopy-height predictions against field measurements before and after removing calendar-date means. Points are colored by crop and compared with identity lines.}
  \label{fig:sup_height_scatter}
\end{figure*}

The two numerically leading methods retain high date-demeaned agreement: Scaffold-GS obtains $R^2=0.972$ and Splatfacto obtains $R^2=0.969$ (Fig.~\ref{fig:sup_height_scatter}). Their within-scene correlations are more moderate, at $r=0.496$ (scene-bootstrap 95\% confidence interval (CI) [0.378, 0.617]) and $r=0.525$ [0.333, 0.699], respectively. Evidence is therefore strongest for point-level height recovery and same-date crop-plot separation, with more moderate support for fine within-scene ranking.

Oat has the highest reconstruction MAE for every method. Because growth stages were not recorded on a standard scale, we report MAE by crop rather than by growth stage.

\begin{table*}[!t]
  \caption{Scene-balanced canopy-height MAE shown separately by crop. Scene counts and sampling-point counts are shown in the headings.}
  \label{tab:sup_height_crop}
  \centering
  \small
  \setlength{\tabcolsep}{8pt}
  \begin{tabular}{@{}lrrrr@{}}
    \toprule
    Method & Wheat (6; 60) & Oat (6; 36) & Corn (11; 66) & Soybean (8; 48) \\
    \midrule
    Nerfacto             & 0.110 & \tabsecond{0.181} & 0.126 & 0.136 \\
    Instant-NGP          & 0.103 & 0.458 & 0.196 & 0.073 \\
    Splatfacto           & 0.063 & 0.192 & \tabbest{0.077} & \tabbest{0.059} \\
    Splatfacto-big       & \tabsecond{0.059} & 0.220 & 0.137 & 0.206 \\
    Mip-Splatting        & 0.069 & 0.280 & 0.207 & 0.092 \\
    Scaffold-GS          & \tabbest{0.050} & \tabbest{0.177} & \tabsecond{0.084} & \tabsecond{0.065} \\
    CityGaussian         & 0.068 & 0.265 & 0.239 & 0.090 \\
    \bottomrule
  \end{tabular}
\end{table*}

\noindent\textbf{Canopy-height sensitivity to the revised export.}
Only exported depth differs in this matched comparison. All seven methods produce finite estimates at all 210 sampling points under both exports.

\begin{table*}[!t]
  \caption{Native-depth $\rightarrow$ revised-depth canopy-height sensitivity for the seven representative methods on the same 31 scenes and 210 sampling points. MAE and RMSE pool the 210 paired predictions.}
  \label{tab:sup_height_configs}
  \centering
  \small
  \setlength{\tabcolsep}{8pt}
  \begin{tabular}{@{}lcc@{}}
    \toprule
    Method & Point MAE (m) & Point RMSE (m) \\
    \midrule
    Nerfacto           & 0.133 $\rightarrow$ 0.133 & 0.158 $\rightarrow$ 0.159 \\
    Instant-NGP        & 0.186 $\rightarrow$ 0.186 & 0.260 $\rightarrow$ 0.256 \\
    Splatfacto         & 0.350 $\rightarrow$ 0.089 & 0.420 $\rightarrow$ 0.118 \\
    Splatfacto-big     & 0.217 $\rightarrow$ 0.145 & 0.346 $\rightarrow$ 0.295 \\
    Mip-Splatting      & 0.169 $\rightarrow$ 0.154 & 0.216 $\rightarrow$ 0.195 \\
    Scaffold-GS        & 0.082 $\rightarrow$ 0.086 & 0.112 $\rightarrow$ 0.113 \\
    CityGaussian       & 0.202 $\rightarrow$ 0.160 & 0.262 $\rightarrow$ 0.199 \\
    \bottomrule
  \end{tabular}
\end{table*}

The largest downstream gain is for Splatfacto. Splatfacto-big and CityGaussian also improve, and Mip-Splatting improves more modestly. Nerfacto, Scaffold-GS, and Instant-NGP change little. The revised export therefore has method-specific downstream effects.

\begin{table*}[!t]
  \caption{Canopy-height validation at complementary aggregation levels over the 210 field measurements. MAE and RMSE in this table pool point-level predictions; the primary table instead macro-averages scene errors. Date-demeaned $R^2$ removes the common calendar-date mean. Within-scene $r$ removes each crop--date--plot scene mean.}
  \label{tab:sup_height_levels}
  \centering
  \small
  \setlength{\tabcolsep}{5pt}
  \begin{tabular}{@{}lrrrrr@{}}
    \toprule
    Method & Point MAE (m) $\downarrow$ & Point RMSE (m) $\downarrow$ & Point $R^2$ $\uparrow$ & Date-dem. $R^2$ $\uparrow$ & Within-scene $r$ $\uparrow$ \\
    \midrule
    Nerfacto             & 0.133 & 0.159 & 0.960 & 0.931 & 0.056 \\
    Instant-NGP          & 0.186 & 0.256 & 0.897 & 0.737 & 0.293 \\
    Splatfacto           & \tabsecond{0.089} & \tabsecond{0.118} & \tabsecond{0.978} & \tabsecond{0.969} & \tabbest{0.525} \\
    Splatfacto-big       & 0.145 & 0.295 & 0.863 & 0.661 & 0.047 \\
    Mip-Splatting        & 0.154 & 0.196 & 0.940 & 0.935 & 0.421 \\
    Scaffold-GS          & \tabbest{0.086} & \tabbest{0.113} & \tabbest{0.980} & \tabbest{0.972} & \tabsecond{0.496} \\
    CityGaussian         & 0.160 & 0.199 & 0.937 & 0.941 & 0.394 \\
    \bottomrule
  \end{tabular}
\end{table*}

\subsection{Effective-LAI trajectories and exploratory scene associations}
\label{sup:lai_scene_quality}

\begin{figure*}[!t]
  \centering
  \includegraphics[width=0.99\textwidth]{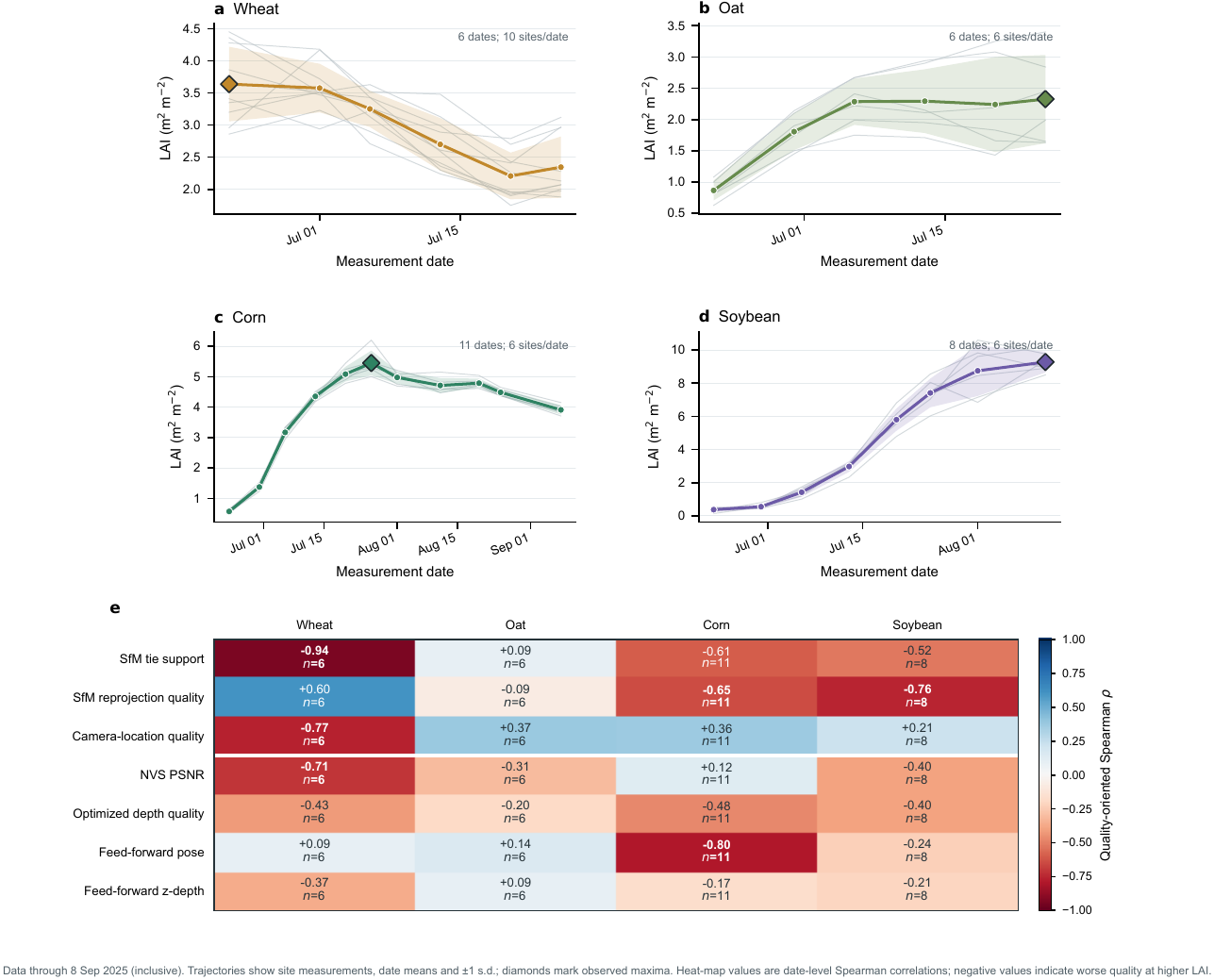}
  \caption{Effective-LAI trajectories and exploratory associations with scene quality in 2025. Panels (a--d) show site measurements, date means with one standard deviation, and the observed maximum for wheat, oat, corn, and soybean. Panel (e) reports date-level Spearman correlations after orienting every quality indicator so that higher values are better; negative values therefore indicate worse quality at higher LAI. Bold type denotes $|\rho|\geq0.65$; all 28 Benjamini--Hochberg-adjusted $q$ values exceed 0.05.}
  \Description{Four crop-specific time-series panels show effective leaf area index through September 8, 2025. A heat map below reports exploratory correlations between LAI and six reconstruction-quality indicators for each crop.}
  \label{fig:sup_lai_scene_quality}
\end{figure*}

The 2025 field subset contains 210 effective-LAI measurements across 31 crop--date scenes. Figure~\ref{fig:sup_lai_scene_quality} includes measurements through September 8 and UAV acquisitions matched within one day. The trajectories capture contrasting crop-development patterns. Higher LAI often coincides with weaker SfM tie support, but the direction and magnitude vary by crop. These date-level correlations are exploratory, and all 28 Benjamini--Hochberg-adjusted $q$ values exceed 0.05.

\subsection{Feed-forward results grouped by crop}
\label{sup:feedforward_crop_results}

Table~\ref{tab:sup_feedforward_crop} reports the Track B metrics after grouping field sequences by crop.

\begin{table*}[!t]
  \caption{Track B results grouped by crop. Scene metrics are first averaged within each field sequence; entries then average four corn sequences, two soybean sequences, and one wheat and oat sequence each.}
  \label{tab:sup_feedforward_crop}
  \centering
  \small
  \setlength{\tabcolsep}{5pt}
  \begin{tabular}{@{}llrrrrr@{}}
    \toprule
    Method & Crop & Pose AUC@5 (\%) $\uparrow$ & $z$-depth AbsRel $\downarrow$ & Point-map AbsRel $\downarrow$ & Inlier@1.03 $\uparrow$ & Scale AbsRel $\downarrow$ \\
    \midrule
    MapAnything & Corn    & \tabbest{93.5} & \tabbest{0.035} & \tabbest{0.037} & \tabbest{0.868} & \tabbest{0.029} \\
                & Soybean & \tabbest{94.4} & \tabbest{0.032} & \tabbest{0.038} & \tabbest{0.889} & \tabbest{0.025} \\
                & Wheat   & \tabbest{93.4} & \tabbest{0.031} & \tabbest{0.034} & \tabbest{0.897} & \tabbest{0.025} \\
                & Oat     & \tabbest{94.4} & \tabbest{0.030} & \tabsecond{0.037} & \tabbest{0.890} & \tabbest{0.024} \\
    \addlinespace
    VGGT        & Corn    & 22.2 & 0.084 & 0.119 & 0.564 & 0.967 \\
                & Soybean & 34.8 & 0.077 & 0.060 & 0.743 & 0.959 \\
                & Wheat   & 11.9 & 0.103 & 0.152 & 0.504 & 0.969 \\
                & Oat     & 31.7 & 0.063 & 0.039 & 0.827 & 0.965 \\
    \addlinespace
    Pi3         & Corn    & \tabsecond{31.6} & \tabsecond{0.068} & \tabsecond{0.116} & \tabsecond{0.579} & 0.965 \\
                & Soybean & \tabsecond{55.3} & \tabsecond{0.034} & \tabsecond{0.043} & \tabsecond{0.845} & 0.957 \\
                & Wheat   & 29.0 & \tabsecond{0.060} & \tabsecond{0.083} & \tabsecond{0.666} & 0.967 \\
                & Oat     & 55.5 & \tabsecond{0.039} & \tabbest{0.035} & 0.873 & 0.961 \\
    \addlinespace
    MASt3R      & Corn    & 26.0 & 0.213 & 0.330 & 0.387 & \tabsecond{0.896} \\
                & Soybean & 47.7 & 0.075 & 0.073 & 0.789 & \tabsecond{0.879} \\
                & Wheat   & \tabsecond{32.5} & 0.127 & 0.144 & 0.617 & \tabsecond{0.884} \\
                & Oat     & \tabsecond{56.7} & 0.052 & 0.047 & \tabsecond{0.882} & \tabsecond{0.890} \\
    \bottomrule
  \end{tabular}
\end{table*}

MapAnything varies little across the sampled crops on all five outputs. The remaining models show distinct failure patterns: VGGT is weakest on wheat for pose and point maps, while MASt3R is weakest on corn for aligned geometry.

\subsection{Paired uncertainty of method ordering}
\label{sup:rank_uncertainty}

Table~\ref{tab:sup_rank_uncertainty} compares the numerical best and runner-up for every main-table metric using 20,000 paired bootstrap replicates. NVS and depth resample the 91 scene pairs. Canopy height resamples the 31 acquisition-matched scenes and recomputes the scene-macro statistic or pooled point-level $R^2$; finite point-level predictions remain nested within scene. Feed-forward evaluation resamples paired scenes within each of the eight sequences and recomputes the equal-sequence macro-average, preserving the hierarchy of the main table. The reported advantage is direction-aligned, so a positive value favors the numerical leader for both higher-is-better and lower-is-better metrics. Win rate is the fraction of original paired scenes on which the leader is better, with ties counted as one half. For height $R^2$, the scene win rate uses lower point-level squared error within the scene because $R^2$ is defined over the pooled sample.

The intervals support the NVS, depth, and feed-forward numerical leaders; all three canopy-height intervals include zero. Individual-scene win rate can differ from the macro ordering because sequence weights and paired-effect magnitudes also matter.

\begin{table*}[!t]
  \caption{Paired uncertainty for every numerical top-versus-runner-up comparison in the main tables. Advantage is oriented so that positive values favor the first method. CI is the paired percentile-bootstrap 95\% interval.}
  \label{tab:sup_rank_uncertainty}
  \centering
  \small
  \setlength{\tabcolsep}{4pt}
  \begin{tabular}{@{}lllrrr@{}}
    \toprule
    Task and sample & Metric & Numerical comparison & Advantage & 95\% CI & Win (\%) \\
    \midrule
    NVS ($n=91$) & PSNR & Splatfacto-big--Splatfacto & 0.35 & [0.31, 0.40] & 90.1 \\
    & SSIM & Splatfacto-big--Splatfacto & 0.051 & [0.047, 0.056] & 95.6 \\
    & LPIPS & Splatfacto-big--Splatfacto & 0.085 & [0.079, 0.090] & 97.8 \\
    & FPS & Splatfacto--Mip-Splatting & 1.41 & [0.80, 1.99] & 68.1 \\
    \addlinespace
    Depth ($n=91$) & RMSE (m) & Scaffold-GS--CityGaussian & 0.127 & [0.109, 0.149] & 100.0 \\
    & AbsRel & Scaffold-GS--CityGaussian & 0.0074 & [0.0067, 0.0082] & 100.0 \\
    & SILog & Scaffold-GS--CityGaussian & 0.0031 & [0.0025, 0.0039] & 98.9 \\
    & Pearson $r$ & Scaffold-GS--CityGaussian & 0.0077 & [0.0055, 0.0101] & 79.7 \\
    \addlinespace
    Height ($n=31$; 210 points) & MAE (m) & Scaffold-GS--Splatfacto & 0.001 & [$-0.011$, 0.012] & 61.3 \\
    & RMSE (m) & Scaffold-GS--Splatfacto & 0.001 & [$-0.012$, 0.012] & 51.6 \\
    & Point $R^2$ & Scaffold-GS--Splatfacto & 0.002 & [$-0.003$, 0.009] & 51.6 \\
    \addlinespace
    Feed-forward ($n=91$; 8 seq.) & Scale AbsRel & MapAnything--MASt3R & 0.863 & [0.860, 0.866] & 100.0 \\
    & Point-map AbsRel & MapAnything--Pi3 & 0.0469 & [0.0379, 0.0564] & 78.0 \\
    & Point-map inlier & MapAnything--Pi3 & 0.186 & [0.169, 0.204] & 86.8 \\
    & ATE RMSE & MapAnything--Pi3 & 0.0443 & [0.0385, 0.0507] & 100.0 \\
    & Pose AUC@5 & MapAnything--Pi3 & 53.64 & [51.98, 55.32] & 100.0 \\
    & $z$-depth AbsRel & MapAnything--Pi3 & 0.0217 & [0.0187, 0.0248] & 89.0 \\
    & $z$-depth inlier & MapAnything--Pi3 & 0.0268 & [0.0118, 0.0421] & 49.5 \\
    & Ray error (\textdegree{}) & Pi3--VGGT & 0.215 & [0.140, 0.294] & 65.9 \\
    \bottomrule
  \end{tabular}
\end{table*}

\subsection{Complete benchmark inventory and scene-level quality control}
\label{sup:scene_inventory}

Table~\ref{tab:sup_scene_inventory} provides a scene-resolved audit of the 91-scene benchmark. Scene IDs follow the release directory structure. The Day aliases are within-year acquisition identifiers, counted from the first survey of that season, and intentionally omit calendar dates. GSD and quality-control indicators are transcribed from the matched processing reports. Registration is the percentage of images successfully oriented, tie points are the triangulated sparse points from bundle adjustment, reprojection is the RMS image reprojection error, and the camera-location residual is the RMS difference between RTK-recorded and bundle-adjusted camera centers.

\begin{table*}[!t]
  \caption{Complete 91-scene benchmark inventory and scene-level photogrammetric quality-control indicators.}
  \label{tab:sup_scene_inventory}
  \centering
  \small
  \setlength{\tabcolsep}{6pt}
  \renewcommand{\arraystretch}{1.08}
  \begin{tabular}{@{}lllrrrrrr@{}}
  \toprule
  Scene ID & Day & Crop & Images &
  \shortstack{GSD\\(mm px$^{-1}$)} &
  \shortstack{Reg.\\(\%)} &
  \shortstack{Tie pts.\\($10^3$)} &
  \shortstack{Reproj.\\(px)} &
  \shortstack{Cam. resid.\\(cm)} \\
  \midrule
  2023/Day001\_Corn & Day1 & Corn & 843 & 3.83 & 100.00 & 233.5 & 1.43 & 2.18 \\
  2023/Day006\_Corn & Day6 & Corn & 840 & 3.83 & 100.00 & 204.7 & 1.51 & 1.98 \\
  2023/Day006\_Soy & Day6 & Soybean & 846 & 3.79 & 100.00 & 211.1 & 1.19 & 2.42 \\
  2023/Day011\_Corn & Day11 & Corn & 839 & 3.75 & 100.00 & 219.9 & 1.62 & 2.12 \\
  2023/Day011\_Soy & Day11 & Soybean & 680 & 3.76 & 100.00 & 168.5 & 1.45 & 3.01 \\
  2023/Day016\_Corn & Day16 & Corn & 843 & 3.65 & 100.00 & 304.5 & 1.35 & 2.73 \\
  2023/Day016\_Soy & Day16 & Soybean & 690 & 3.67 & 100.00 & 163.1 & 1.13 & 2.43 \\
  2023/Day020\_Corn & Day20 & Corn & 842 & 3.69 & 100.00 & 348.8 & 1.50 & 1.80 \\
  2023/Day020\_Soy & Day20 & Soybean & 688 & 3.71 & 100.00 & 194.9 & 1.36 & 2.67 \\
  2023/Day026\_Corn & Day26 & Corn & 839 & 3.66 & 100.00 & 352.1 & 1.39 & 2.23 \\
  2023/Day026\_Soy & Day26 & Soybean & 688 & 3.73 & 100.00 & 200.3 & 1.37 & 2.32 \\
  2023/Day031\_Corn & Day31 & Corn & 850 & 3.58 & 100.00 & 351.2 & 1.22 & 1.81 \\
  2023/Day031\_Soy & Day31 & Soybean & 693 & 3.71 & 100.00 & 200.2 & 1.22 & 2.01 \\
  2023/Day037\_Corn & Day37 & Corn & 847 & 3.64 & 100.00 & 356.3 & 1.33 & 1.77 \\
  2023/Day037\_Soy & Day37 & Soybean & 690 & 3.72 & 100.00 & 227.3 & 1.36 & 1.85 \\
  2023/Day042\_Corn & Day42 & Corn & 851 & 3.66 & 100.00 & 355.6 & 1.36 & 2.19 \\
  2023/Day042\_Soy & Day42 & Soybean & 691 & 3.69 & 100.00 & 195.6 & 1.44 & 2.08 \\
  2023/Day047\_Corn & Day47 & Corn & 975 & 3.65 & 100.00 & 435.8 & 1.27 & 2.51 \\
  2023/Day047\_Soy & Day47 & Soybean & 703 & 3.74 & 100.00 & 244.1 & 1.36 & 2.52 \\
  2023/Day052\_Corn & Day52 & Corn & 837 & 3.75 & 100.00 & 389.7 & 1.48 & 1.90 \\
  2023/Day052\_Soy & Day52 & Soybean & 670 & 3.77 & 100.00 & 352.5 & 1.63 & 3.00 \\
  2024/Day001\_Corn1 & Day1 & Corn & 1,285 & 3.96 & 100.00 & 283.6 & 0.97 & 5.78 \\
  2024/Day001\_Corn2 & Day1 & Corn & 1,391 & 4.03 & 100.00 & 266.4 & 0.93 & 2.25 \\
  2024/Day015\_Corn1 & Day15 & Corn & 1,279 & 3.92 & 100.00 & 452.1 & 1.12 & 4.66 \\
  2024/Day015\_Corn2 & Day15 & Corn & 1,372 & 3.99 & 100.00 & 678.3 & 1.16 & 2.50 \\
  2024/Day023\_Corn1 & Day23 & Corn & 1,180 & 3.91 & 100.00 & 430.3 & 0.95 & 2.14 \\
  2024/Day023\_Corn2 & Day23 & Corn & 1,164 & 3.86 & 100.00 & 579.1 & 1.19 & 2.85 \\
  2024/Day036\_Corn1 & Day36 & Corn & 574 & 3.79 & 100.00 & 684.5 & 1.16 & 2.15 \\
  2024/Day036\_Corn2 & Day36 & Corn & 489 & 3.58 & 96.93 & 665.3 & 0.94 & 6.05 \\
  2024/Day045\_Corn1 & Day45 & Corn & 1,048 & 4.29 & 100.00 & 1,411.1 & 1.17 & 2.33 \\
  2024/Day045\_Corn2 & Day45 & Corn & 745 & 4.89 & 99.87 & 878.4 & 1.18 & 3.69 \\
  2024/Day056\_Corn1 & Day56 & Corn & 806 & 4.80 & 100.00 & 693.9 & 1.47 & 43.50 \\
  2024/Day056\_Corn2 & Day56 & Corn & 663 & 4.78 & 98.79 & 646.5 & 1.37 & 3.64 \\
  2024/Day074\_Corn1 & Day74 & Corn & 565 & 5.68 & 100.00 & 514.1 & 1.60 & 2.70 \\
  2024/Day074\_Corn2 & Day74 & Corn & 512 & 5.71 & 100.00 & 543.3 & 1.61 & 4.40 \\
  2024/Day099\_Corn1 & Day99 & Corn & 809 & 4.72 & 100.00 & 617.5 & 1.22 & 2.48 \\
  2024/Day099\_Corn2 & Day99 & Corn & 516 & 5.70 & 100.00 & 504.1 & 1.21 & 2.67 \\
  2024/Day120\_Corn1 & Day120 & Corn & 799 & 4.78 & 100.00 & 656.9 & 1.18 & 3.21 \\
  2024/Day120\_Corn2 & Day120 & Corn & 515 & 5.80 & 100.00 & 445.5 & 1.14 & 3.16 \\
  2025/Day001\_Oat & Day1 & Oat & 700 & 4.79 & 100.00 & 254.9 & 1.09 & 2.62 \\
  2025/Day001\_Wheat & Day1 & Wheat & 908 & 4.67 & 100.00 & 329.5 & 1.31 & 2.01 \\
  2025/Day007\_Corn & Day7 & Corn & 1,210 & 5.84 & 100.00 & 265.0 & 1.12 & 4.72 \\
  2025/Day007\_Oat & Day7 & Oat & 807 & 4.94 & 100.00 & 276.4 & 1.17 & 2.80 \\
  2025/Day007\_soybean & Day7 & Soybean & 1,081 & 4.88 & 100.00 & 196.3 & 0.96 & 2.34 \\
  2025/Day007\_Wheat & Day7 & Wheat & 1,669 & 4.49 & 100.00 & 381.5 & 1.42 & 2.48 \\
  2025/Day012\_Oat & Day12 & Oat & 689 & 4.94 & 100.00 & 236.4 & 1.15 & 2.14 \\
  \bottomrule
  \end{tabular}
\end{table*}

\begin{table*}[!t]
  \ContinuedFloat
  \caption{Complete 91-scene benchmark inventory and scene-level photogrammetric quality-control indicators (continued).}
  \label{tab:sup_scene_inventory_cont}
  \centering
  \small
  \setlength{\tabcolsep}{6pt}
  \renewcommand{\arraystretch}{1.08}
  \begin{tabular}{@{}lllrrrrrr@{}}
  \toprule
  Scene ID & Day & Crop & Images &
  \shortstack{GSD\\(mm px$^{-1}$)} &
  \shortstack{Reg.\\(\%)} &
  \shortstack{Tie pts.\\($10^3$)} &
  \shortstack{Reproj.\\(px)} &
  \shortstack{Cam. resid.\\(cm)} \\
  \midrule
  2025/Day012\_Wheat & Day12 & Wheat & 1,419 & 4.45 & 100.00 & 517.4 & 1.41 & 1.87 \\
  2025/Day019\_Corn & Day19 & Corn & 1,200 & 5.76 & 100.00 & 268.5 & 1.14 & 5.27 \\
  2025/Day019\_Oat & Day19 & Oat & 691 & 4.97 & 100.00 & 304.7 & 1.51 & 3.17 \\
  2025/Day019\_soybean & Day19 & Soybean & 1,070 & 4.84 & 100.00 & 189.8 & 1.06 & 3.17 \\
  2025/Day019\_Wheat & Day19 & Wheat & 1,398 & 4.56 & 100.00 & 887.5 & 1.69 & 2.67 \\
  2025/Day021\_Oat & Day21 & Oat & 694 & 4.94 & 100.00 & 270.6 & 1.32 & 2.03 \\
  2025/Day021\_Wheat & Day21 & Wheat & 1,423 & 4.39 & 100.00 & 676.4 & 1.43 & 2.14 \\
  2025/Day024\_Oat & Day24 & Oat & 691 & 4.92 & 100.00 & 299.6 & 1.51 & 2.26 \\
  2025/Day024\_Wheat & Day24 & Wheat & 1,419 & 4.41 & 100.00 & 582.4 & 1.39 & 1.81 \\
  2025/Day028\_Corn & Day28 & Corn & 1,204 & 5.80 & 100.00 & 321.3 & 1.36 & 3.27 \\
  2025/Day028\_Oat & Day28 & Oat & 679 & 4.97 & 100.00 & 413.1 & 1.81 & 2.55 \\
  2025/Day028\_soybean & Day28 & Soybean & 1,085 & 4.86 & 100.00 & 279.0 & 1.03 & 3.03 \\
  2025/Day028\_Wheat & Day28 & Wheat & 1,402 & 4.50 & 100.00 & 872.4 & 1.72 & 2.73 \\
  2025/Day030\_Oat & Day30 & Oat & 676 & 4.98 & 100.00 & 420.8 & 1.68 & 2.78 \\
  2025/Day030\_Wheat & Day30 & Wheat & 1,411 & 4.52 & 100.00 & 903.8 & 1.66 & 2.97 \\
  2025/Day033\_Corn & Day33 & Corn & 1,170 & 5.77 & 100.00 & 445.1 & 1.50 & 2.19 \\
  2025/Day033\_Oat & Day33 & Oat & 675 & 5.02 & 100.00 & 484.1 & 1.66 & 2.34 \\
  2025/Day033\_soybean & Day33 & Soybean & 1,087 & 4.85 & 100.00 & 337.9 & 1.06 & 2.06 \\
  2025/Day033\_Wheat & Day33 & Wheat & 1,418 & 4.54 & 100.00 & 909.9 & 1.59 & 2.06 \\
  2025/Day036\_Oat & Day36 & Oat & 658 & 4.99 & 100.00 & 373.3 & 1.72 & 2.24 \\
  2025/Day036\_Wheat & Day36 & Wheat & 1,422 & 4.53 & 100.00 & 613.4 & 1.58 & 1.97 \\
  2025/Day040\_Corn & Day40 & Corn & 1,207 & 5.72 & 100.00 & 725.5 & 1.54 & 2.78 \\
  2025/Day040\_Oat & Day40 & Oat & 651 & 5.03 & 100.00 & 390.2 & 1.74 & 2.28 \\
  2025/Day040\_soybean & Day40 & Soybean & 1,082 & 4.88 & 100.00 & 536.6 & 1.26 & 2.23 \\
  2025/Day040\_Wheat & Day40 & Wheat & 1,410 & 4.59 & 100.00 & 759.4 & 1.89 & 2.52 \\
  2025/Day044\_Oat & Day44 & Oat & 671 & 4.94 & 100.00 & 337.4 & 1.96 & 2.65 \\
  2025/Day044\_Wheat & Day44 & Wheat & 1,421 & 4.47 & 100.00 & 446.0 & 1.82 & 2.95 \\
  2025/Day047\_Corn & Day47 & Corn & 1,204 & 5.76 & 100.00 & 834.2 & 1.45 & 2.70 \\
  2025/Day047\_Oat & Day47 & Oat & 683 & 5.03 & 100.00 & 282.9 & 1.77 & 2.29 \\
  2025/Day047\_soybean & Day47 & Soybean & 1,084 & 4.90 & 100.00 & 414.4 & 1.34 & 2.08 \\
  2025/Day047\_Wheat & Day47 & Wheat & 1,415 & 4.52 & 100.00 & 539.9 & 1.72 & 2.00 \\
  2025/Day052\_Corn & Day52 & Corn & 1,207 & 5.65 & 100.00 & 609.7 & 1.98 & 2.53 \\
  2025/Day052\_Oat & Day52 & Oat & 678 & 5.00 & 100.00 & 276.4 & 1.76 & 2.88 \\
  2025/Day052\_soybean & Day52 & Soybean & 1,082 & 4.87 & 100.00 & 377.1 & 1.46 & 2.34 \\
  2025/Day052\_Wheat & Day52 & Wheat & 1,415 & 4.51 & 100.00 & 524.6 & 1.84 & 2.34 \\
  2025/Day059\_Corn & Day59 & Corn & 1,204 & 5.47 & 100.00 & 687.0 & 1.72 & 2.16 \\
  2025/Day059\_soybean & Day59 & Soybean & 1,083 & 4.84 & 100.00 & 382.9 & 1.28 & 2.25 \\
  2025/Day069\_Corn & Day69 & Corn & 1,259 & 5.48 & 100.00 & 703.9 & 1.67 & 2.12 \\
  2025/Day069\_soybean & Day69 & Soybean & 1,086 & 4.84 & 100.00 & 368.6 & 1.26 & 2.48 \\
  2025/Day077\_Corn & Day77 & Corn & 1,205 & 5.54 & 100.00 & 800.5 & 1.55 & 2.22 \\
  2025/Day077\_soybean & Day77 & Soybean & 1,087 & 4.87 & 100.00 & 510.7 & 1.23 & 2.10 \\
  2025/Day083\_Corn & Day83 & Corn & 1,206 & 5.68 & 100.00 & 747.6 & 1.56 & 2.08 \\
  2025/Day083\_soybean & Day83 & Soybean & 1,081 & 4.88 & 100.00 & 573.2 & 1.33 & 2.10 \\
  2025/Day096\_Corn & Day96 & Corn & 1,207 & 5.52 & 100.00 & 708.4 & 1.57 & 2.22 \\
  2025/Day096\_soybean & Day96 & Soybean & 1,089 & 4.87 & 100.00 & 380.8 & 1.23 & 2.02 \\
  \bottomrule
  \end{tabular}
\end{table*}

\end{document}